\documentclass{new_tlp}
\submitted{31 March 2017}
\revised{31 December 2017}
\accepted{14 June 2018}
\usepackage{CDMO-DLP-TPLP2017}
\usepackage[framemethod=TikZ]{mdframed} 
\usepackage{multirow}

\title[Logic Programming as a Service]{Logic Programming as a Service}
\author[R. Calegari et al.]
	{ROBERTA CALEGARI, ENRICO DENTI\\
		{\sc Alma Mater Studiorum}--Universit\`a di Bologna, Italy\\\email{roberta.calegari@unibo.it}, \email{enrico.denti@unibo.it}
	\and STEFANO MARIANI\\
		Universit\`{a} di Modena e Reggio Emilia, Italy\\\email{stefano.mariani@unimore.it}
	\and ANDREA OMICINI\\
		{\sc Alma Mater Studiorum}--Universit\`a di Bologna, Italy\\\email{andrea.omicini@unibo.it}}

\begin{document}
\maketitle
\begin{abstract}
New generations of distributed systems are opening novel perspectives for logic programming (LP): on the one hand, service-oriented architectures represent nowadays the standard approach for distributed systems engineering; on the other hand, pervasive systems mandate for situated intelligence.
In this paper we introduce the notion of \emph{Logic Programming as a Service} (\lpaas{}) as a means to address the needs of pervasive intelligent systems through logic engines exploited as a distributed service. 
First we define the abstract architectural model by re-interpreting classical LP notions in the new context; then we elaborate on the nature of LP interpreted as a service by describing the basic \lpaas{} interface.
Finally, we show how \lpaas{} works in practice by discussing its implementation in terms of distributed \tuprolog{} engines, accounting for basic issues such as interoperability and configurability.
\end{abstract}
\begin{keywords}
	logic programming,
	distributed systems,
	service-oriented architecture,
	pervasive systems,
	intelligent systems,
	\lpaas{},
	situatedness
\end{keywords}

\section{Introduction}
\labelsec{Intro}
Computation is moving towards pervasive, ubiquitous environments where devices, software agents, and services are expected to seamlessly integrate and cooperate in support of human users, anticipating their needs and more generally acting on their behalf, delivering services in an ``anywhere, anytime'' fashion \cite{mobiledevagents-ciaV,sapere-pmc10years}.
Even more, software agents, robots, sensors, etc.\ could work together with people for a common goal, with the same level of efficiency and expertise as human-only teams.
Such systems could face important challenges in several fields---from military network-centric operations, to gaming technologies, simulation, computer security, transportation and logistics, and others \cite{distributedintelligence-jpa2}.

The above scenarios naturally fit a distributed approach: tasks are often distributed in space, time, or functionality, and their completion can clearly benefit from the chance of solving subproblems modularly and concurrently.
At the same time, the same scenarios inherently call for intelligence -- namely, \emph{distributed situated intelligence} \cite{distributedintelligence-jpa2} -- to exploit domain knowledge, understand the local context, and share information in support of intelligent applications and services \cite{pervasiveontology-ker18,situatedintelligence-mam2017}.

\emph{Logic programming} (LP henceforth) boasts a long-respected reputation in supporting intelligence: originally conceived for single solvers and later extended towards concurrency and parallelism, LP has the potential to fully support pervasive computing scenarios once it is suitably \emph{re-interpreted}.
Re-interpretation of LP should develop along three main lines:
\textit{(i)} \emph{architecture}---that is, the need to go beyond the (originally monolithic) structure of LP systems, which is unsuitable for distributed contexts such as \iot{} mobility/cloud ecosystems, typically grounded upon the service-oriented computing paradigm \cite{soabook-2005};
\textit{(ii)} \emph{situatedness}---that is, enabling logic theories, queries, and resolutions to be context-aware w.r.t.\ the (computational) environment, space, and time;
\textit{(iii)} \emph{interaction}---that is, the opportunity to re-think the interaction patterns used by clients to query logic engines, which should lean towards on-demand computation.

At the same time, LP declarativeness and explicit knowledge representation enable knowledge sharing at the most adequate level of abstraction, while supporting modularity and separation of concerns \cite{incrementalreasoning-ccis190}, which are specially valuable in open and dynamic distributed systems (\emph{serendipitous interoperability}, \citeNP{serendipitousinteroperability-jaise5}).
As a further element, LP soundness and completeness straightforwardly enable agents' intelligent reasoning.
Finally, LP extensions or logic-based computational models -- such as meta-reasoning about situations \cite{situatedlp-ker19} or labelled variables systems \cite{labelledlp-fi161} -- could be incorporated so as to enable complex behaviours tailored to the situated components.

Although LP languages and technologies represent in principle a natural candidate for injecting intelligence within computational systems \cite{brownlee2011}, and despite the many practical application developed over the years -- see \citeN{Palu:2010-LPapp}, \citeN{martelli1995} for a survey --, the adoption of LP in pervasive contexts has been historically hindered by technological obstacles -- efficiency, integration issues -- as well as by some cultural resistance towards LP-based approaches outside the academy. 
However, technology advancements, on the one hand, and the emergence of the \iot{} context, on the other, are drastically changing such a scenario, possibly allowing LP to unleash its full potential in real-world applications.

Along this line, in this paper we present \emph{Logic Programming as a Service} (\lpaas{}), a novel approach intended as the natural evolution of distributed LP in pervasive systems, explicitly designed to exploit context-awareness so as to promote the distribution of situated intelligence within smart environments.
As the name suggests, the basic idea is to deliver LP-based intelligence \emph{as a service}, granting ubiquitous access to knowledge and on-demand reasoning via LP services, spread over the network and configured to respond to specific local needs.
Accordingly, some classical LP notions need to be revised and extended: for instance, client/service interaction is no longer bound to the traditional console-based query/response loop, and is instead redesigned to provide the dynamism, flexibility, and expressiveness required by the targeted application scenarios---e.g., \iot{}.
Similarly, time and space situatedness promotes novel forms of client/service interaction, enabling clients to submit ``situated'' queries where the notions of time and locus explicitly affect the computation.

The remainder of the paper is organised as follows.
After \xs{relatedEvolution} reviews the main works about the evolution of distributed LP, \xs{vision} introduces the vision behind \lpaas{}, by discussing how the service perspective and the new situated dimension of computation mandate for a re-interpretation of some basic LP concepts.
\xs{architecture} shows how such a re-interpretation affects LP at the architectural level, by discussing more practically the logic-based service-oriented architecture supporting \lpaas{}.
\xs{service} defines the \lpaas{} service interface, and elaborates on the interaction patterns.
\xs{casestudy} presents a prototype implementation developed on the top of the \tuprolog{} system, 
while \xs{discussion} discusses a case study in the Smart House field. Related works are reported in \xs{related}.

\section{Distributed LP: Evolution}
\labelsec{relatedEvolution}
Research on distributed intelligence has gained increasing popularity over the years \cite{distributedintelligence-jpa2}.
Starting from the seminal work of \citeN{parallellp-fpca1981}, concurrency, parallelism, and several approaches for distributing intelligence have been explored---from LP languages specifically designed for distribution, to pure logic-based models, rule-based systems, probabilistic graphical models, and ontologies.
In the following we organise and describe some of the most relevant contributions to the field and to our approach, motivating the need for further advancement.

\paragraph{Implicit Parallelism.}
The first efforts to advance beyond sequential LP start from the programming schemes for the interpretation of logic programs---in particular, towards implicit parallel evaluation, leading to explore AND-parallelism, OR-parallelism, Search parallelism, and Stream-AND-parallelism.

\citeN{naf-logicdbbook1978} introduces a scheme that allows negative literals in queries; some years later, the Naish scheme \cite{nuprolog-iclp1988} introduces co-routing among procedure calls.
Meanwhile, \citeN{resolution-iclp1984} focus on AND-parallel evaluation: their asynchronous version corresponds to the execution models of parallel LP languages.
These schemes perform and adapt well to different forms of parallelism: however, they are not meant to face distributed programming.
Also, it is worth noting that implicit parallelism lacks two important control mechanisms: synchronisation of logic processes, and control over the non-determinism of schedulers.

\paragraph{Explicit Parallelism.}
Later approaches focus on ``extraction'' of parallelism via explicit language constructs.

A first line of research moves from concurrent logic languages, rooted in the Relational Language \cite{parallellp-fpca1981}, generally acknowledged as the first concurrent LP language.
In \concurrentprolog{} \cite{concurrentprolog-book1}, Guarded Horn Clauses \cite{guardedhornclauses-lp1985}, and Parlog \cite{parlog-parle1987}, goal evaluation is carried out by a network of fine-grained logic processes (i.e., atomic goals) that are executed in parallel: processes communicate via shared streams, i.e., bi-directional channels on which data items flow.

An alternative research line follows the idea of extending \prolog{} with special features for distributed execution, like message passing.
This approach preserves the operational semantics of sequential \prolog{}, augmenting the language with ad-hoc communication primitives.
One of the major references in this field is \deltaprolog{} \cite{deltaprologsemantics-tapsoft1989}, where \prolog{} is extended with constructs for sequential and parallel composition of goals, inter-process communication and synchronisation, and external non-determinism.
\deltaprolog{} programs \cite{deltaprolog-iclp1989} using concurrency mechanisms do not lend themselves to the usual declarative interpretation as Horn clauses, and are grounded instead on the theory of Distributed Logic \cite{dlplpgic-implprologbook1984}.
This approach extends Horn clause logic with the notion of time-dependent events, on which process communication and synchronisation are based, making distributed logic a special kind of temporal logic.

Besides enabling inter-process communication for logic programs, orthogonal aspects such as their deployment are not considered, neither the issues brought along by distribution -- such as validity in time of logic theories and their global consistency -- are taken into account.

\paragraph{Agents, Communication, and Coordination for Distributed LP.}
Further steps towards distributed LP come with \sharedprolog{} \cite{sharedprolog-toplas13}, based on parallel \emph{agents} that are \prolog{} programs extended with a guard mechanism.
The programmer controls the granularity of parallelism, coordinating agents' communication and synchronisation via a centralised data structure, the \emph{blackboard}, inspired to the model defined in \cite{blackboard-aimagazine7} as well as to the \linda{} coordination model \cite{linda-toplas7}.
The main idea is to exploit the blackboard within the logic framework to coordinate logic processes.
However, the inference engine is not situated in time and space, i.e. the query result is independent from the entities' position, the time flow, and context/situation changes.

\paragraph{LP in Pervasive, Context-aware Systems.}
More recently, LP has been explored as a promising solution to bring intelligence into pervasive context-aware systems.

\citeN{folcontextaware-puc7} show that using first-order logic is a very effective and powerful way of dealing with context, promoting an approach to develop a flexible and expressive model supporting context-awareness, enabling deduction of higher-level situations from perceptions about basic contexts---via rule-based approaches.
A key advantage of formally modelling the context is that the expressiveness of the model itself can be clearly specified and automatically verified.
\citeN{situatedlp-ker19} emphasises that LP is generally useful for context reasoning, as well as for supporting rule-based (meta)programming in context-aware applications, enabling, i.e. hierarchical description of complex situations in terms of other situations.
This approach encourages a high level of abstraction for representing and reasoning about situations, and supports building context-aware systems incrementally through modularity and separation of concerns.
The focus on context-awareness of both contributions is at the base of our choice of re-interpreting distributed LP by targeting especially context-aware systems, as pervasive ones usually are---being the \iot{} a prominent example.

Other works take different approaches: from pure logic-based models, to rule-based systems and probabilistic graphical models, up to ontologies.

Rule-based systems \cite{context-chi1999,context-puc5,rulebased-icps2006,context-sagaware2011} have been in use for decades for both model representation and reasoning in context-aware applications.
More recently, \citeN{Nalepa2014} have proposed a rule-based, \textit{learning} middleware for storage and reasoning in a distributed scenario.
The idea is to delegate context acquisition to middleware, that is, a rule-based context reasoning platform tailored to the needs of intelligent distributed mobile computing devices.
The need for a dedicated middleware layer is apparent in the aforementioned works, further strengthening the idea that distributed LP is not confined to context manipulation, and deserves instead general attention.

In \citeN{ranganathan:2004}, fuzzy and probabilistic logic is exploited to handle the uncertainty of the environment and deal with the imperfections of data.
Probabilistic graphical models \cite{context-pmc6} can be exploited to support the modelling of, and the reasoning about, uncertain information in pervasive systems, even if exact inference in complex probabilistic models can be a NP-hard task.
Description logic, usually used in combination with ontologies, is another LP extension effective for modelling concepts, roles, individuals, and their relationships, as well as to provide simple reasoning capabilities \cite{dl4pervasive-ccis315}.
However, only simple classification tasks can be solved, and no mechanisms are provided to infer more complex information from existing data.
Also, design and implementation are typically more difficult and time-consuming than with other approaches.
Since uncertainty of information is the natural enemy of global consistency, our approach moves from the choice of abandoning the idea of globally-consistent (in terms of both time and space) logic theories (or, knowledge bases---KB) in favour of \emph{locally}-consistent ones.

\section{The \lpaas{} Vision}
\labelsec{vision}

The evolution of LP in parallel, concurrent, and distributed scenarios is the main motivation for re-interpreting the notion of \emph{distribution} of LP in today's context.
Since Service-Oriented Architecture (\soa{}) is the \emph{de facto} standard for distributed application development in both the academia and the industry \cite{soabook-2005}, \xss{service} focuses on how LP can be re-interpreted in the \emph{service} perspective.
This perspective further emphasises the role of \emph{situatedness}, already brought along by distribution in itself: thus, \xss{situatedness} discusses how being situated in space, time, and context affects LP computation.
The two novel perspectives are merged together in \xss{towards-situated-service}, which develops the idea of LP as a \emph{situated service}.

\subsection{The Service Perspective}
\labelssec{service}

The service-oriented perspective deeply affects the way in which LP engines are conceived, designed, and used---in particular, as far as the very nature of LP \emph{encapsulation} is concerned, the way in which clients interact (requiring \emph{statelessness}), and the assumptions about the surrounding context (\emph{locality}) are concerned.

\paragraph{Encapsulation.}
A service hides both data representation and the computational mechanisms behind a public interface exposed to its clients.
In the context of LP engines, this means that both the logic theory (the data) and the resolution process (the computational mechanism) are \emph{inaccessible} -- and, in general, \emph{not observable} -- from outside the boundary of the service interface.
As a consequence, theory manipulation mechanisms, such as \texttt{assert}/\texttt{retract}, are no longer directly applicable from the client perspective: since the logic theory is encapsulated by the service, dedicated mechanisms are required for its handling.
For instance, in an \iot{} scenario, this may happen via a separate ``sensor API'' through which sensor devices regularly update the KB of the LP service according to their perception of the surrounding environment.

Accordingly, the logic theory of a \lpaas{} service can be either \emph{static} or \emph{dynamic} (which are mutually exclusive configurations).
The way in which the LP service can be accessed obviously depends on that: time is an issue for dynamic KB, not for static ones.

\paragraph{Statelessness.}
Encapsulation makes it irrelevant \emph{how} the encapsulated behaviour is implemented: what actually matters are the inputs triggering, and the outputs resulting from, that behaviour.
Furthermore, in the \soa{} perspective, services are usually redundantly distributed over a network of hosts for enhancing the service availability and reliability: thus, it does not really matter \emph{who} actually carries out the encapsulated behaviour.
In the context of LP, this means that interactions with clients should be also allowed to be \emph{stateless}---that is, include all the information required by the resolution process, since a different component may serve a different request.
Notably, statelessness is the default for RESTful web services, too.

It is worth emphasising here that statelessness does not contrast with the above encapsulation property, since the former regards the \emph{invocation} of \lpaas{} services -- hence the interaction between clients and servers --, whereas the latter concerns \lpaas{} services themselves---that is, their inner nature.
In other words, statelessness implies that servers are not supposed to track the state of interactions, so that a service request should not assume or rely on previous interactions, whereas encapsulation means that only the selected properties of the service are visible and modifiable from the outside.

At the same time, in order to cope with data-intensive applications, where stateless interaction may become cumbersome, \lpaas{} also supports \emph{stateful} interaction---yet, at the clients' convenience and will.
This is particularly handy for scenarios where reasoning and inference should be based on continuous and possibly unbounded streams of data, such as those coming from sensors in \iot{} deployments.

\paragraph{Locality.}
The distributed nature of the system drastically changes the perspective over consistency:
maintaining \emph{globally-consistent} information is typically unfeasible in such systems.
Furthermore, when pervasive systems enter the picture, even \emph{globally-available} information is usually not a realistic assumption: for instance, in \iot{} scenarios, heterogeneous data streams are continuously made available by sensor devices scattered in specific portions of the physical environment.
As a consequence, encapsulation is inevitably bound to a specific, (\emph{local}) portion of the system---with a notion of locality extending up to when/where availability and/or consistency are inevitably lost.

In the context of LP, this means first of all resorting to a \emph{multi-theory} logical framework, exploiting the typical approach to \emph{modularity} adopted in traditional LP in order to allow for parallel and concurrent computation \cite{bugliesi-jlp19}.
Also, locality implies that each logic theory describes just what is \emph{locally} true ---which basically means leaving aside in principle the global acceptation of the \emph{closed world} assumption \cite{closedworld-logicdbbook1978} in favour of a more realistic \emph{locally-closed world} assumption.
Accordingly, every LP service is to be queried about \emph{what is locally known to be true}, with no need to resort to global knowledge of any sort---and with no need to distribute the resolution process in any way.

\subsection{The Situatedness Perspective}
\labelssec{situatedness}

The distribution of LP service instances directly calls for \emph{situatedness}, intended as the property of the LP service to be immersed in the surrounding computational/physical environment, whose changes may affect its computations \cite{coordarch-eaai41}.
As an example, new sensor data may change the replies of an LP service to queries.
Situatedness adds three new dimensions to LP computations: \emph{space}, \emph{time}, and \emph{context}.

\paragraph{Space.}
To be situated in space means that the \emph{spatial context} where the LP service is executing may affect its properties, computations, and the way it interacts with clients.

Distribution \emph{per se} constitutes a premise to spatial situatedness: each LP instance runs on a different device, thus on a different network host, therefore accessing the different computational and network resources that are \emph{locally} available.
Moreover, since LP services encapsulate the logic theory for their resolution process, the locally-gathered knowledge affects the result, once it is represented in terms of logic axions.
		
Also, more articulated forms of spatial situatedness may be envisioned: for instance, \emph{mobile} clients may request LP services from different locations at each request, possibly even \emph{while} moving, which means that the LP service must be able to coherently identify and track clients so as to reply to the correct network address.
Finally, it is possible in principle to conceive logic theories -- or even individual axioms therein -- with spatially-bound validity, that is, that are true only in specific points or regions in space---analogously to \emph{spatial tuples} in \citeN{spatialtuples-idc2016}.

\paragraph{Time.}
Complementarily, being situated in time means that the \emph{temporal context} when the LP service is executed may affect its properties, computations, and interactions with clients.
Yet again, distribution alone already brings about temporal issues: moving information in a network takes time, thus aspects such as expiration of requests, obsolescence of logic theories, and timeliness of replies should be taken into account when designing the LP service.

Furthermore, since reconstructing a \emph{global} notion of time in pervasive systems is either unfeasible or non-trivial, each LP service should operate on its own local time---that is, computing deadlines, leasing times, and the like according to its \emph{local perception} of time.
Also, in the same way as for spatial situatedness, temporal situatedness may also imply that logic theories or individual axioms may have their time-bounded validity---e.g., holding true up to a certain instant in time, and no longer since then.

\paragraph{Context.}
Besides the space/time fabric, situatedness also regards the generic \emph{environment} within which LP services execute---that is, the computational and physical context which may affect their working cycle: for instance, it may depend on the available CPUs and RAM, whether an accelerometer is available on the current hosting device, etc.
		
A basic level of \emph{contextual situatedness} is already embedded in the very nature of the LP service: in fact, locality of the resolution process implies that the logic theory for goal resolution belongs to the context of the LP service, thus straightforwardly affecting its behaviour.
However, especially in the \iot{} scenarios envisioned for \lpaas{}, the computational and physical contexts may both impact the LP service: e.g., sensor devices may continuously update the service KB with their latest perceptions, while actuators may promptly provide feedback on success/failure of physical actions.

\subsection{Towards LP as a Situated Service}
\labelssec{towards-situated-service}

The above perspectives promote a radical re-interpretation of a few facets of LP, moving LP itself towards the notion of \lpaas{} envisioned in this paper---that is, in terms of a \emph{situated service}.
Such a notion articulates along four major lines:
\begin{itemize}
  \item the preservation (with re-contextualisation) of the SLD resolution process;
  \item stateless interactions;
  \item time-sensitive computations;
  \item space-sensitive computations.
\end{itemize}

\paragraph{The re-contextualisation of the SLD resolution process.}
The SLD resolution process \cite{resolution-jacm12} remains a staple in \lpaas{}: yet, it is re-contextualised in the situated nature of the specific LP service.
This means that, given the precise \emph{spatial}, \emph{temporal}, and \emph{general} contexts within which the service is operating \emph{when the resolution process starts}, the  process follows the usual rules of SLD resolution: situatedness is accounted for through the service abstraction with respect to such three contexts.

With respect to the \emph{spatial context}, the resolution process obviously takes place in the hosting device \emph{where} the LP service is running, thus taking into account the specific properties of the computational and physical environment therein available -- CPU, RAM, network facilities, GPS positioning, etc.\ -- there included the specific logic theory the LP service relies on.
As mentioned in \xss{situatedness}, more articulated forms of spatial situatedness -- e.g., involving mobility of clients (and LP services, in principle), or, virtual/physical regions of validity for logic axioms -- could be envisioned.

The \emph{temporal context} refers to the resolution process taking place on a \emph{frozen snapshot} of the LP service state -- there including its KB --, which stays unaffected to external stimuli (possibly affecting the resolution process) until the process itself terminates.
This way, despite the dynamic nature of the KB -- encapsulated by the service abstraction -- which could change e.g. due to sensors' perceptions, the resolution process is guaranteed to operate on a consistent stable state of the logic theory.

Finally, the resolution process depends on the \emph{general context} of the specific device hosting the LP service instance---thus considering the state of the KB therein available, as assembled by e.g., the set of sensors devices therein available, the service agents gathering new local information, and so on.

\paragraph{Stateless interactions.}
A first change brought by \lpaas{} concerns interaction with clients of the LP service.

In classical LP, interactions are necessarily \emph{stateful}: the user first sets the logic theory, then defines the goal, and then asks for one or more solutions, iteratively.
This implies that the LP engine is expected to store the logic theory to exploit as its KB, to memorise the goal under demonstration, and to track how many solutions have been already provided to the user---and all these items become part of the state of the LP engine.

Instead, in \lpaas{} interactions are first of all (even though not exclusively) \emph{stateless}: coherently with \soa{}, the LP service instance that actually serves each request may be different at each time, e.g.\ due to redundancy of distributed software components aimed at improving availability and reliability of the LP service.
In such a perspective, each client query (interaction) should be possibly self-contained, so that it does not matter which specific service instance responds---because there is no need for it to track the state of the interaction session.

It is worth emphasising that in the case of stateful interaction, adequate measures need be taken to prevent possible problems related to different \lpaas{} service instances serving repeated requests, requests from mobile clients, and similar situations.
Two possible solutions could be considered for this concern: 
\textit{(i)} the \lpaas{} middleware could simply \emph{lock} a service in case of stateful interaction, ensuring that the client always interacts with the same service instance (this is essentially the problem of \emph{conversational continuity}, well documented in the literature \cite{linda-toplas7}); 
\textit{(ii)} alternatively, the \lpaas{} middleware could take care of the \emph{hand-off} of the interaction from instance to instance, ensuring proper sharing of the information needed to preserve statefulness across service instances.

\paragraph{Time-local computation.}
Another change stemming from the situated nature of \lpaas{} is concerned with the relationship between the resolution process and the flow of time.

In pure LP, the logic theory is simply assumed to be always valid, and time-related aspects  do not affect the resolution process; for instance, assertion / retraction mechanisms are most typically regarded as extra-logic ones.
As discussed above, in \lpaas{} the consistency of the resolution process is guaranteed by the fact that the possibly ever-changing KB encapsulated by the service is \emph{frozen} in time when the resolution process itself begins: nevertheless, time situatedness requires by definition that time affects the LP service computation in some way.

Accordingly, in \lpaas{} each axiom in the KB is decorated with a \emph{time interval}, indicating the time validity of each clause.
Every time a new resolution process starts in order to serve a \lpaas{} request, the logic theory used is the one containing all and only the axioms holding true at the \emph{timestamp} associated with the resolution process itself.
In the simplest case, such a timestamp is implicitly assigned by the LP server as the current local time when the request for goal demonstration is first served.
Otherwise, it could also be explicitly assigned by clients along with the request---e.g., defining a specific time when asking for a goal demonstration.

\paragraph{Space-local computation.}
Analogously, classical LP has no notion of space situatedness: be it a virtual or a physical space, the LP engine is a monolithic component providing its ``services'' only locally, to its co-located ``clients'' working on the same machine.

The \lpaas{} interpretation stems again from the very nature of service in modern \soa{}-based applications---a computational unit providing its functionalities through a network-reachable endpoint.
Therefore, the resolution process in \lpaas{} is naturally and inherently affected by the specific \emph{computational locus} where a given LP service instance is executing at a given moment---there including the locally-available resources.

\section{The \lpaas{} Architecture}
\labelsec{architecture}

\lpaas{} is a logic-based, service-oriented approach for \emph{distributed situated intelligence}, conceived and designed as the natural evolution of LP in nowadays pervasive computing systems.
Its purpose is to enable situated reasoning via explicit definition of the spatio-temporal structure of the environment where situated entities act and interact.

Along the lines traced in \xss{towards-situated-service}, we now elaborate more practically on how encapsulation, statelessness, and locality -- that is, the \emph{service perspective} (\xss{service}) -- are exploited in \lpaas{} according to the three dimensions of \emph{situatedness} described in \xss{situatedness}---that is, time, space, and context.
Then, we briefly describe microservices  \cite{microservices-saasbook} as a key enabler architecture for \lpaas{}.

\subsection{Service Architecture}
\labelssec{servicearchitecture}

\paragraph{Encapsulation.}
As it straightforwardly stems from \soa{} principles, encapsulation is exploited in \lpaas{} so as to define a standard API that shields \lpaas{} clients from the inner details of the service while providing suitable means of interaction.

\begin{figure}
	\includegraphics[width=0.475\linewidth]{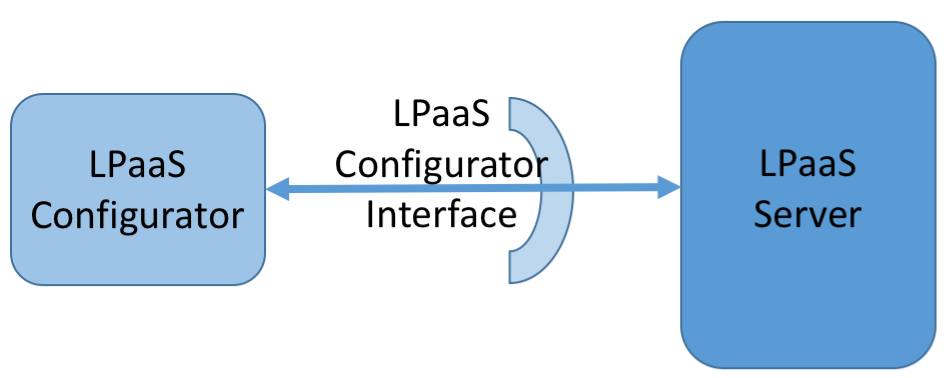}
	\includegraphics[width=0.475\linewidth]{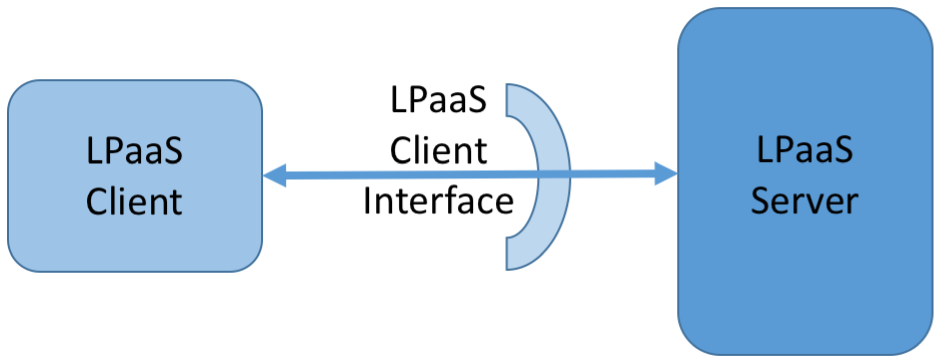}
	\caption{\lpaas{} Configurator Service Architecture (left) and Client Service Architecture (right)}\labelfig{service}
\end{figure}

Accordingly, each LP server node exposes its LP service to clients via two interfaces, depicted in \xf{service}:
\begin{description}
	\item[Client Interface] exposes methods for \emph{observation} and \emph{usage}.
		\textit{Client} refers to any kind of users, either individuals (humans, software agents) or groups entitled to exploit the \lpaas{} services.
	\item[Configurator Interface] enables service \emph{configuration} and requires proper access credentials.
		\textit{Configurator} refers to service managers---privileged agents with the right of enforcing control and policies for that local portion of the system.
\end{description}
Applications can access the service as either \textit{Clients} or \textit{Configurators}, via the corresponding interfaces.
The service is initialised at deployment-time on the server machine: once started, it can be dynamically re-configured at run-time by any configurator.

\paragraph{Locality.}

Situatedness is exploited as a means to consistently handle \emph{locality} w.r.t.\ context, time, and space.

In fact, dealing with situated logic theories means to give up with the idea of global consistency in a closed world: in \lpaas{} multiple KB are spread throughout a network infrastructure, likely geographically distributed, executing within different computational contexts, and possibly either fed by sensors or manipulated by service agents perceiving the physical context.
By allowing distributed access and reasoning over its own locally-situated knowledge base, each \lpaas{} node actively contributes to the overall availability of the global knowledge.

Accordingly, pervasive application scenarios where logic theories represent local knowledge inherently call for \emph{dynamic KB}, \emph{autonomously} evolving during the service lifetime\footnote{Here,``autonomously'' means that in the \lpaas{} perspective the logic KB may evolve over time with no need for a client to invoke \texttt{assert/retract}, or equivalent methods -- which, in fact, are not included in the \lpaas{} standard API detailed in \xss{api} -- but, e.g., due to sensor devices' perceptions transparently feeding the LP service KB.}.
As such, each situated KB of a \lpaas{} service can be seen as representing what is known to be true and relevant in a given location in space at a given time, thus possibly changing over time -- e.g., due to data streams coming from sensor devices --, and potentially different from any other KB located elsewhere---as depicted in \xf{sit-arch}.

\begin{figure}
	\includegraphics[width=0.60\linewidth]{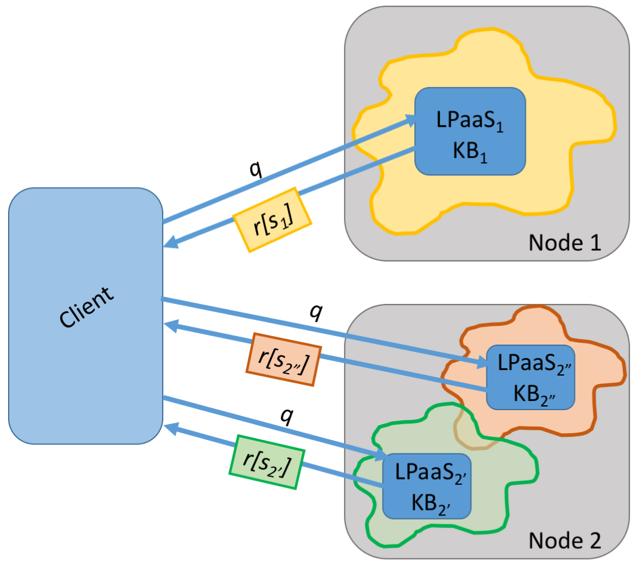}
	\caption{Situatedness of \lpaas{}: the same query ($q$) by the same client may be resolved differently ($\mathit{r}[s_{1}], \mathit{r}[s_{2'}], \mathit{r}[s_{2''}]$) by distinct \lpaas{} services ($\mathit{LPaaS}_{1}, \mathit{LPaaS}_{2'}, \mathit{LPaaS}_{2''}$) based on their local computational, physical, and spatio-temporal context ($\mathit{S}_{1}, \mathit{S}_{2'}, \mathit{S}_{2''}$)}
	\labelfig{sit-arch}
\end{figure}

Accordingly,
\begin{itemize}
	\item each \lpaas{} clause has a lifetime, expressed as a time interval of validity---as in the case of the current temperature in a room;
	\item as a result, at any point $t$ in time a \lpaas{} service has precisely one logic theory made of all and only the clauses that hold true at time $t$;
	\item each \lpaas{} resolution process is either implicitly (by the \lpaas{} server) or explicitly (by the \lpaas{} client) labelled with a \emph{timestamp}, used to determine the KB to be used for the resolution itself---which then works as the standard LP resolution.
\end{itemize}

\paragraph{Statelessness.}
\emph{Uncoupling} is one of the main requirements for interaction in distributed systems: that is why \lpaas{} provides stateless client-server interaction as one of its main features.
The same holds true in particular for pervasive systems, where \emph{instability} is one of the main issues, as well as for mobile systems, with any sort of mobility: physical mobility of users and devices; users who change their computing device while using applications; service instances migrating from machine to machine---as in a cloud-based environment.

The need for uncoupling promotes \emph{stateless interaction} in \lpaas{}.
Thus, for instance, both \lpaas{} clients and service instances can freely move with no concerns for requests tracking and identity/location bookeeping.

In order to balance the effect of statelessness on data-intensive interactions between LP service and users, \lpaas{} also provides clients with the ability to ask for more than one solution at a time, and even all of them, with a single request.
Nevertheless, \lpaas{} also makes it possible to obtain a \emph{stream of solutions} from the resolution process, rather than a single solution at a time in an individual interaction session, to better meet the needs of fast-paced dynamic scenarios in which clients want to be constantly updated by the LP service about some situation.

Accordingly, \lpaas{} provides clients with the means to obtain both stateless and stateful client-server interaction:
\begin{description}
	\item[stateful] once the logic theory to consider is settled, and the goal stated, the client should be able to ask for any amount of solutions -- possibly iteratively, possibly at different times and from different places -- with the service being responsible to guarantee consistency and validity of solutions by keeping track of the related interaction sessions with the same client;
	\item[stateless] in this case, no session state is tracked by the server, so each client request should contain all the information required to serve the request itself.
\end{description}
It is worth highlighting that nothing prevents the service from being stateful and stateless simultaneously, because the LP server can manage multiple kinds of requests concurrently; instead, of course, each client request in \lpaas{} is either stateful or stateless.

\subsection{{M}icroservices as Technology Enablers}
\labelssec{microservices}
Service-oriented architectures represent nowadays the standard approach for distributed system engineering \cite{soabook-2005}: so, \lpaas{} adopts the \emph{Software as a Service} (\saas{}) architecture as its reference \cite{saas-cacm53}.

Accordingly, information technology resources are conceived as continuously-provided services:
\saas{} applications are supposed to be available 24/7, scale up \& down elastically, support resiliency to changes (i.e., in the form of suitable fault-tolerance mechanisms), provide a responsive user experience on all popular devices, and require neither user installation nor application updates.

In particular, LP services in \lpaas{} can be fruitfully interpreted as \emph{microservices} \cite{microservices-saasbook}.
Microservices are a recent architectural style for \saas{} applications promoting usage of self-contained units of functionally with loosely-coupled dependencies on other services: as such, they can be designed, developed, tested, and released independently.
Thanks to their features, microservices are deserving increasing attention also in the industry -- pretty much like \soa{} in the mid 2000s --  where fast and easy deployment, fine-grained scalability, modularity, and overall agility are particularly appreciated \cite{microservicespitfalls-book2016}.

Technically speaking, microservices are designed to expose their functionality through standardised network-addressable APIs and data contracts, making it possible to choose the programming language, operating system, and data store that best fit the service needs and the developers' skills set, without worrying about interoperability.
Microservices should also be dynamically \emph{configurable}, possibly in different forms and with different configuration levels.
Obviously, actual support to interoperability requires multiple levels of standardisation: to this end, \lpaas{} defines its own \emph{interfaces} for both \emph{configuration} and \emph{exploitation}, while relying on widely adopted standards as far as the representation formats (i.e., \citeNP{JSON-home}) and interaction protocols (i.e.\ REST over HTTP, or \citeNP{MQTT-home}) are concerned.

\section{The \lpaas{} Service}
\labelsec{service}

Following the reference architecture above, designing \lpaas{} amounts first of all at defining the Configurator Interface and the Client Interface---as in \xf{service}.

Generally speaking, the LP service should support \emph{(i)} \emph{observational methods} to provide configuration and contextual information about the service, \emph{(ii)} \emph{usage methods} to trigger computations and reasoning, as well as to ask for solutions,
and \emph{(iii)} \emph{configuration methods} to allow the configurator to set the LP service configuration.

Observational methods make it possible to query the service about its configuration (stateful/stateless, static/dynamic), the state of the knowledge base, and the admissible goals: as such, they belong to the Client Interface, but can be made available also in the Configurator Interface for convenience.
Usage methods, instead, belong uniquely to the Client Interface: they allow clients to ask for one or more solutions---one solution, $n$ solutions, or all solutions available, for stateful or stateless requests as well.
Configurator methods belong uniquely to the Configurator Interface, and are intended to set the service configuration, KB nature, and admissible goals.

\subsection{Service Interfaces}
\labelssec{api}

\begin{table}[!tb]
\caption{\lpaas{} Configurator Interface}\label{LPaaS-ConfigInterface}
\begin{tabular}{c}
\hline\hline
\tt setConfiguration(+ConfigurationList)\\
\tt getConfiguration(-ConfigurationList)\\
\tt resetConfiguration()\\
\\
\tt setTheory(+Theory)\\
\tt getTheory(-Theory)\\
\tt setGoals(+GoalList)\\
\tt getGoals(-GoalList)\\
\hline\hline
\end{tabular}
\end{table}

Adopting the \prolog{} notation for input/output \cite{prologstandard-book1996}, the actual Configurator methods are detailed in Table~\ref{LPaaS-ConfigInterface}, while the Client Interface is detailed in Table~\ref{LPaaS-ClientInterface}.
Since the first is rather self-explanatory, we focus on the Client Interface.

The first thing worth noting is that usage predicates for stateless and stateful requests are slightly different from each other.
In the case of stateless requests, the \texttt{solve} operation is conceptually atomic and self-contained---so, e.g., the \texttt{Goal} to solve is always one of its arguments; instead, in the case of stateful requests it is up to the server to keep track of the request state, so the goal is to be set only once by the client before the first \texttt{solve} request is issued.

The second key aspect is the threefold impact of \emph{time awareness}: regardless of whether the server is either computing or idle, time flows anyway, so predicates have to be time-sensitive.
Accordingly,
\begin{itemize}
	\item \texttt{solve} predicates can also contain a \texttt{Timeout} parameter (server time) for the resolution, so as to avoid blocking the server indefinitely: if the resolution process does not complete within the given time, the request is cancelled, and a negative response is returned;
	\item for \emph{stateful requests}, the client could also ask for a \emph{stream} of solutions, which is particularly useful in \iot{} scenarios exploiting sensor devices, or monitoring processes: to this end, \texttt{solve} takes a \texttt{time} argument (server time), meaning that each new solution should be returned not faster than every \texttt{time} milliseconds;
	\item when the KB is \emph{dynamic}, all predicates take an additional \texttt{Timestamp} argument, meaning that each theory has a \emph{time-bounded validity}: this feature can be used during the proof of a goal to ensure that only the clauses valid at the given \texttt{Timestamp} are taken into account in that resolution process.
\end{itemize}
For the sake of convenience, \texttt{solveAfter} methods are provided for mimicking the LP stateful interaction on a stateless request channel, fast-forwarding to the \texttt{N+1} solution \texttt{AfterN}.

Finally, the \texttt{reset} primitive resets the resolution process, with no need to reconfigure the service (i.e., re-select the goal);
in contrast, the \texttt{close} primitive actually closes the communication with the server, so the goal must be re-set before re-querying the server.

\begin{table}
\caption{\lpaas{} Client Interface}
\label{LPaaS-ClientInterface}
\begin{minipage}{\textwidth}\scriptsize
\begin{tabular}{cc}
\hline\hline
\multicolumn{2}{c}{\textbf{STATIC KNOWLEDGE BASE}}		\\
\hline\hline
\textbf{Stateless}	& \textbf{Stateful}\\
\hline
\multicolumn{2}{c}{\tt{getServiceConfiguration(-ConfigList)}}		\\
\multicolumn{2}{c}{\tt getTheory(-Theory)}		\\
\multicolumn{2}{c}{\tt getGoals(-GoalList)} 					\\
\multicolumn{2}{c}{\tt isGoal(+Goal)} 						\\
\\
& \tt setGoal(template(+Template)) 	\\
& \tt setGoal(index(+Index)) 	\\
\\
 \tt solve(+Goal, -Solution)       				& \tt solve(-Solution) 	\\
 \tt solveN(+Goal, +NSol, -SolutionList)       		& \tt solveN(+N, -SolutionList) \\
 \tt solveAll(+Goal, -SolutionList)		       		& \tt solveAll(-SolutionList) 	 \\
 \\
 \tt solve(+Goal, -Solution, within(+Time))       	& \tt solve(-Solution, within(+Time)) 	\\
 \tt solveN(+Goal, +NSol, -SolutionList, within(+Time)) & \tt solveN(+NSol, -SolutionList, within(+Time)) \\
 \tt solveAll(+Goal, -SolutionList, within(+Time))	& \tt solveAll(-SolutionList, within(+Time))	\\
 \\
 \tt solveAfter(+Goal, +AfterN, -Solution)		& 				 \\
 \tt solveNAfter(+Goal, +AfterN, +NSol, -SolutionList)		&  			 \\
 \tt solveAllAfter(+Goal, +AfterN, -SolutionList)	& 			 \\
 \\
 & \tt solve(-Solution, every(@Time))	\\
 & \tt solveN(+N, -SolutionList, every(@Time))	\\
 & \tt solveAll(-SolutionList, every(@Time))	\\
 & \tt pause()	\\
 & \tt resume()	\\
 \\
 \multicolumn{2}{c}{\tt reset()} 					\\
\multicolumn{2}{c}{\tt close()} 						\\
\hline\hline
\multicolumn{2}{c}{\textbf{DYNAMIC KNOWLEDGE BASE}}		\\
\hline\hline
\textbf{Stateless}	& \textbf{Stateful}\\
\hline
\multicolumn{2}{c}{\tt{getServiceConfiguration(-ConfigList)}}		\\
\multicolumn{2}{c}{\tt getTheory(-Theory, ?Timestamp)}		\\
\multicolumn{2}{c}{\tt getGoals(-GoalList)} 					\\
\multicolumn{2}{c}{\tt isGoal(+Goal)} 						\\
\\
& \tt setGoal(template(+Template)) 	\\
& \tt setGoal(index(+Index)) 	\\
 \\
\tt solve(+Goal, -Solution, ?Timestamp)       	& \tt solve(-Solution, ?Timestamp) 	\\
\tt solveN(+Goal, +NSol, -SList, ?TimeStamp)       		& \tt solveN(+N, -SolutionList, ?TimeStamp) \\
\tt solveAll(+Goal, -SList, ?TimeStamp)		       		& \tt solveAll(-SolutionList, ?TimeStamp) 	 \\
\\
\tt solve(+Goal, -Solution, within(+Time), ?TimeStamp)       	& \tt solve(-Solution, within(+Time), ?TimeStamp) 	\\
\tt solveN(+Goal, +NSol, -SList, within(+Time), ?TimeStamp) & \tt solveN(+NSol, -SList, within(+Time), ?TimeStamp) \\
\tt solveAll(+Goal, -SList, within(+Time), ?TimeStamp)	& \tt solveAll(-SList, within(+Time), ?TimeStamp)	\\
\\
\tt solveAfter(+Goal, +AfterN, -Solution, ?TimeStamp)		& 				 \\
\tt solveNAfter(+Goal, +AfterN, +NSol, -SList, ?TimeStamp)		&  			 \\
\tt solveAllAfter(+Goal, +AfterN, -SList, ?TimeStamp)	& 			 \\
\\
& \tt solve(-Solution, every(@Time), ?TimeStamp)	\\
& \tt solveN(+N, -SList, every(@Time), ?TimeStamp)	\\
& \tt solveAll(-SList, every(@Time), ?TimeStamp)	\\
& \tt pause()	\\
& \tt resume()	\\
\\
\multicolumn{2}{c}{\tt reset()} 					\\
\multicolumn{2}{c}{\tt close()} 						\\
\hline\hline
\end{tabular}
\end{minipage}
\end{table}

\subsection{Computational Model}
\labelssec{model}

\begin{figure}
	\includegraphics[width=0.75\linewidth]{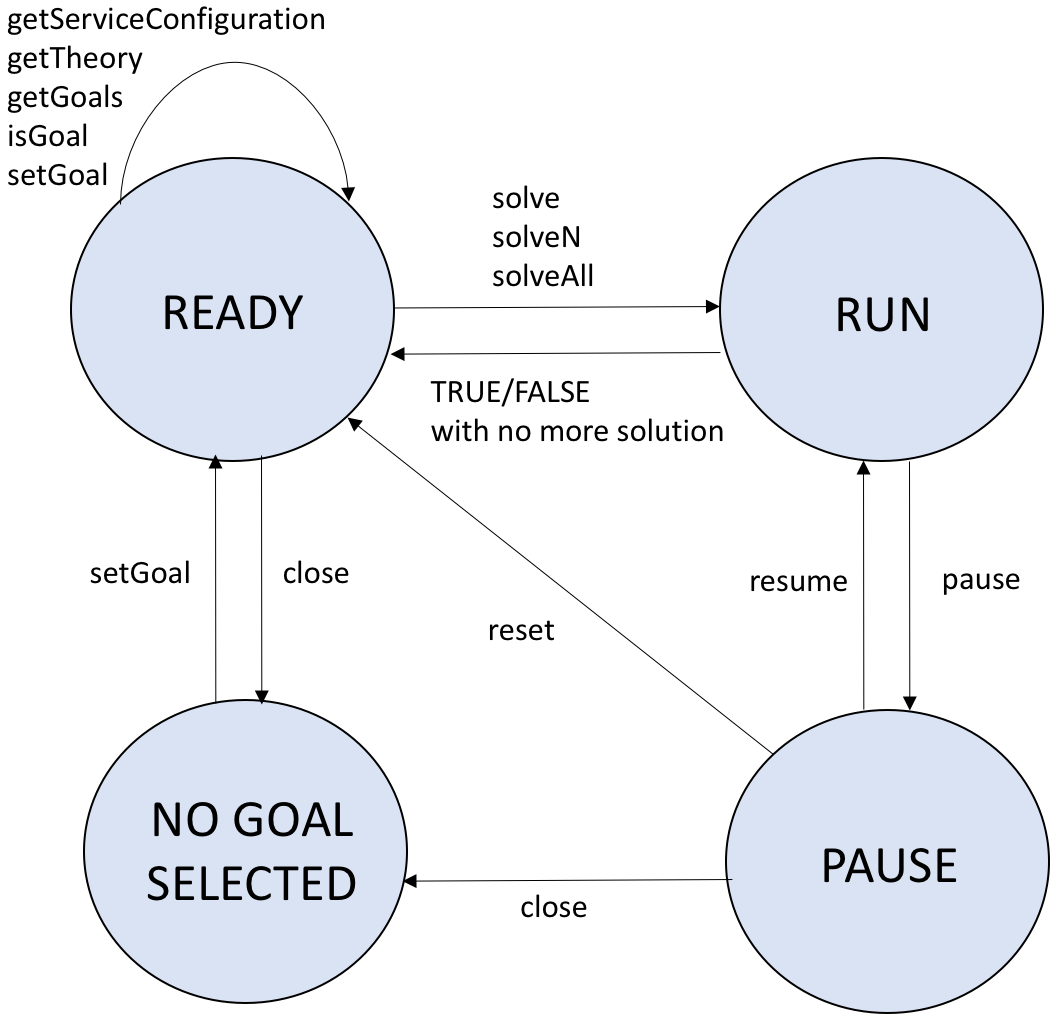}
	\caption{The \lpaas{} Finite State Machine}\labelfig{stateMachine}
\end{figure}

The computational model of the service is depicted by the Finite State Machine (FSM) in \xf{stateMachine}, made of four states:
\begin{itemize}
	\item \emph{ready} (initial state) | where the service is started and the engine is configured;
	\item \emph{run} | where the service is undergoing some resolution process triggered by queries;
	\item \emph{pause} | representing the temporary suspension of computations;
	\item \emph{no goal selected} (final state) | when the client connection is closed.
\end{itemize}
In the \emph{ready} state, the service can be queried about its properties and a new goal can be set, thus defining a new resolution process.
When a new query is submitted, the service moves to the \emph{run} state, indicating that a resolution process is taking place.
Computation may then be paused several times, causing the service to move back and forth from the \emph{pause} state: from there, resolution can also be reset (coming back to the initial state), or closed (moving to state \emph{no goal selected}).

\section{\lpaas{} in \tuprolog{}}
\labelsec{casestudy}
To test the effectiveness of the proposed model and architecture, we implement a first prototype of \lpaas{} as a RESTful Web Service (WS) \cite{webarchtecture-toit2}, embracing the \emph{Web of Things} (\wot{}) vision \cite{wot-computer48}.
Accordingly, our approach follows the \wot{} perspective in re-interpreting the ``things'' (as well as their functionalities and data streams) as RESTful resources accessible through WS protocols, addressing the need to harmonically exploit all the components of the \iot{} system by virtualising individual things in some sort of software abstraction.
There, each interaction session starts with a client request conveying the so-called ``method'' information (i.e.\ how the receiver has to process the request) and the ``scope'' information (i.e.\ which is the target data).
Then, computations occur on the receiving side, where the target resource applies the method to the scope.
The result is a response conveying an optional representation of the requested resource (functionality or data).

The computational model of the prototype reflects the state machine described in \xf{stateMachine}.
We reuse and adapt patterns commonly used for the REST architectural style, and introduce a novel architecture which supports the embedding of \prolog{} engines into WS.
\xf{WSarch} (left) shows the general architecture of the server side and its components (access interfaces, \prolog{} engine, and data store), as well as some exemplary client applications interacting via HTTP requests and JSON objects.

\begin{figure}
	\centering
  	\includegraphics[width=0.54\linewidth]{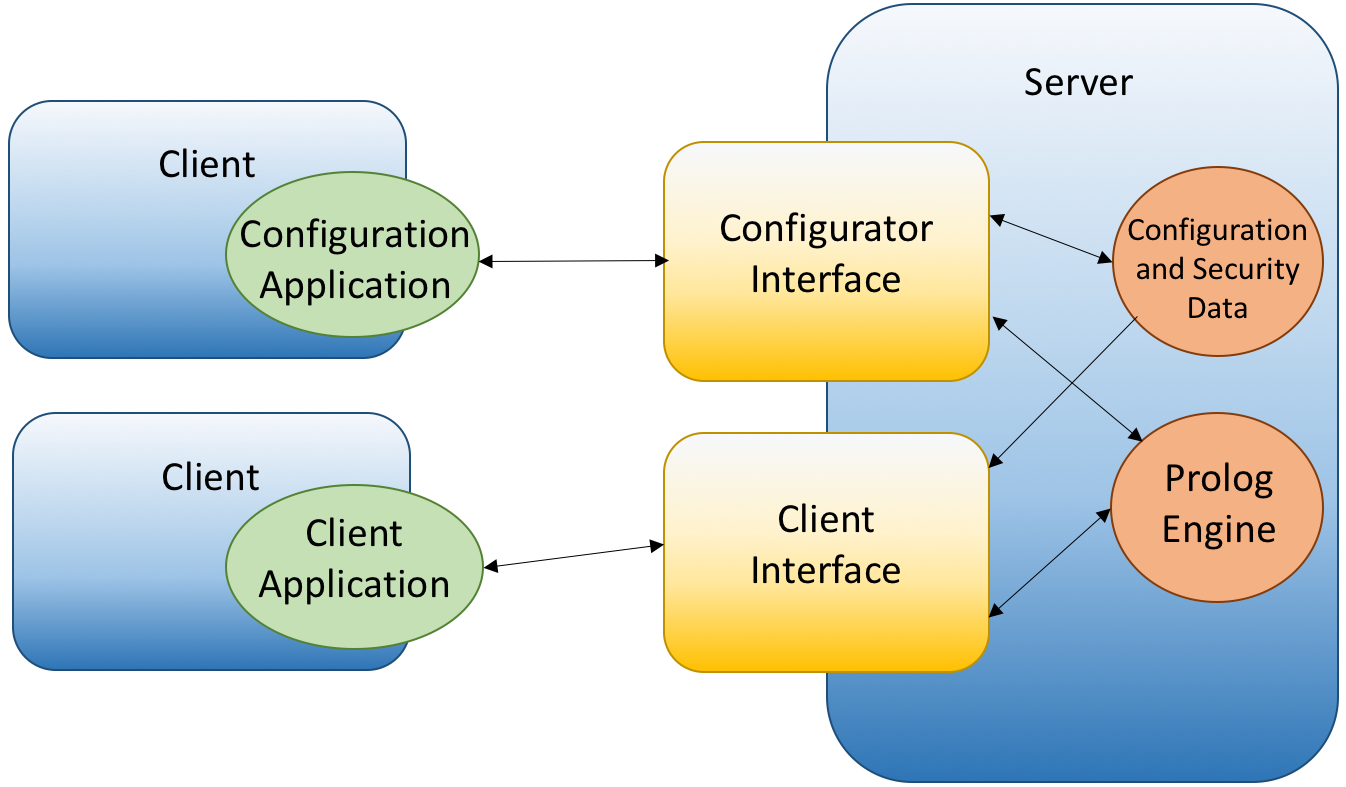}
	\includegraphics[width=0.45\linewidth]{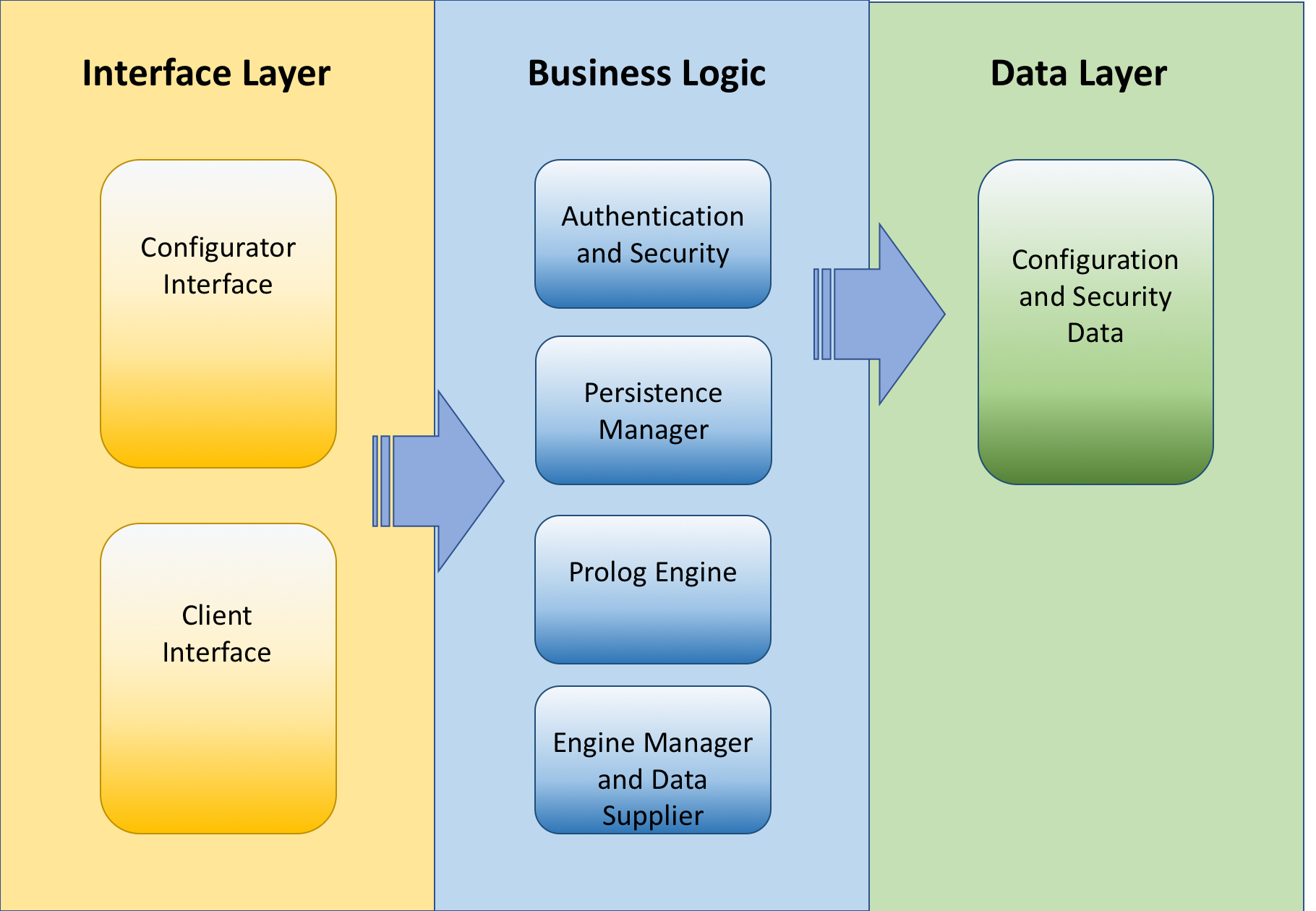}
  	\caption{The \lpaas{} RESTful Web Service (left) and Server architecture (right)}\labelfig{WSarch}
\end{figure}

\xf{WSarch} (right) shows the server inner architecture, composed of three logical layers: the interface, the business logic, and the data layer.
The interface layer encapsulates the Configurator and Client interfaces.
The business logic layer wraps the \prolog{} engine with the aim of managing incoming requests consistently.
The data layer is responsible for managing the data store tracking, i.e., all the configuration options necessary to restore the service in case of unpredictable shutdown (i.e., operating parameters and security metadata such as clients' role, username, password).

Since those data are expected to be rather limited in size for most scenarios, we choose to keep them in the server application so as to offer a light-weight, self-contained service: however, they could be easily moved to a separate persistence layer on, i.e., an external DB application, if necessary.

The server implementation is realised by exploiting a plurality of technologies that are commonly found in the \soa{} field: the business logic is realised on the J2EE framework \cite{J2EE-home}, exploiting EJB \cite{EJB-home}, while the database interaction is implemented on top of JPA \cite{JPA-home}.

\begin{figure}
	\centering
  	\includegraphics[width=\linewidth]{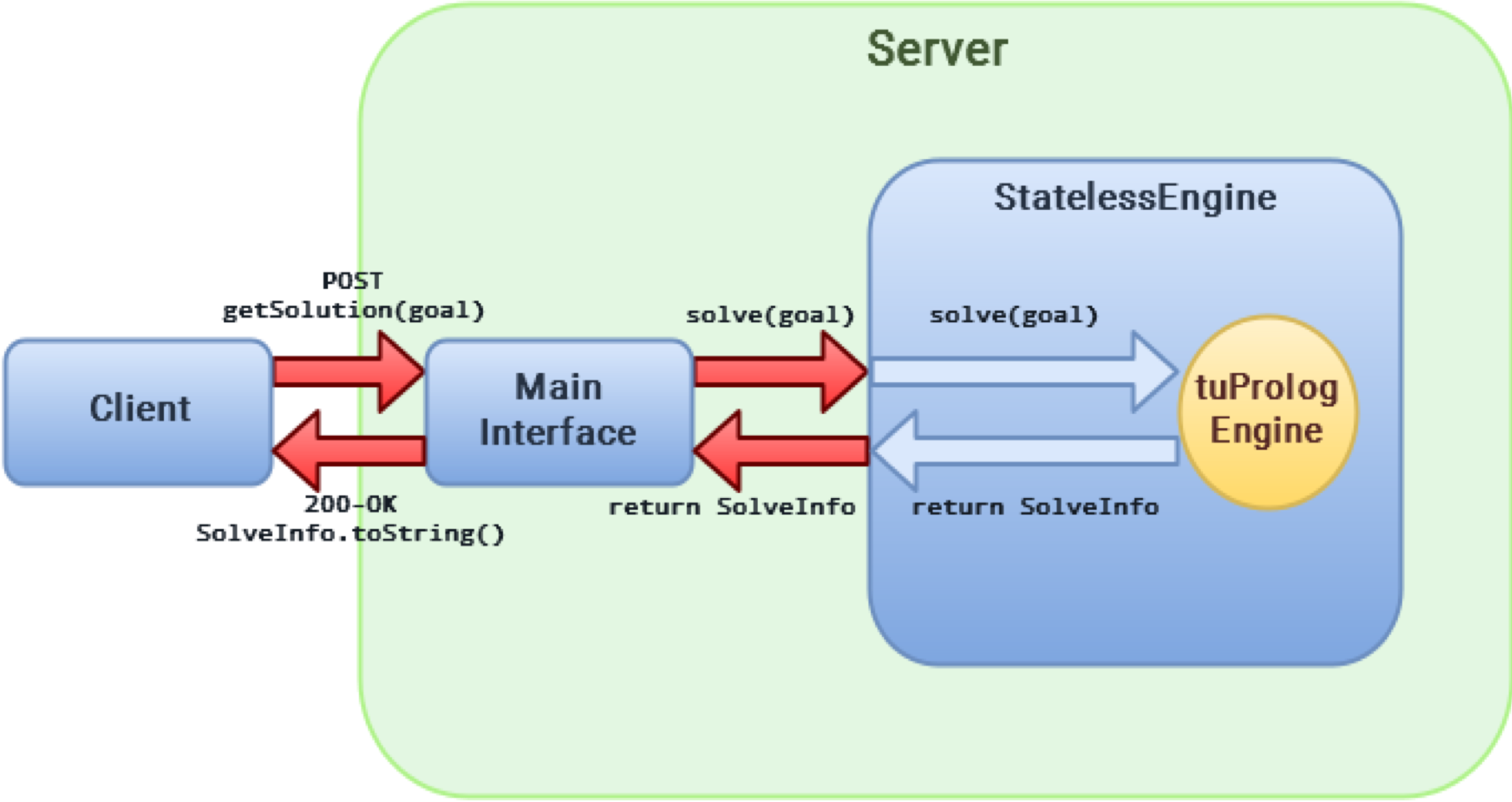}
  	\caption{Client Server Interaction -- Inner calls}\labelfig{ClientServerInteraction}
\end{figure}

The \prolog{} engine is implemented on top of the \tuprolog{} system \cite{tuprolog-padl01}, which provides not only a light-weight engine, particularly well-suited for the envisioned pervasive computing scenarios, but also a multi-paradigm and multi-language working environment, paving the way towards further forms of interaction and expressiveness.
Also, \tuprolog{} 3.2 supports JSON serialisation natively, ensuring the interoperability required by a WS.
The \tuprolog{} engine, distributed as a Java JAR or Microsoft .NET DLL, is easily deployable and exploitable by applications as a \emph{library service}---that is, from a software engineering standpoint, a suitably encapsulated set of related functionalities.

\xf{ClientServerInteraction} shows a client-server interaction in case of a stateless request.
The StatelessEngine component, realised exploiting a Stateless Bean, wraps the \prolog{} engine object to manage the concurrent requests transparently.

The service interfaces exploit the EJB architecture, but can also be accessed as RESTful Web Services, realised using JAX-RS Java Standard implemented in the Jersey library \cite{jersey-home}.
Security is based on JOSE \cite{jose4-home}, an open source (Apache 2.0) implementation of JWT and the JOSE specification suite.
The application is deployed using the Payara Application Server \cite{payara-home}, a Glassfish open-source fork.

\subsection{\tuprolog{}-as-a-Service in action}
\labelssec{2P-aaS}
The \tuprolog{}-as-a-Service prototype is freely available on Bitbucket \cite{lpaas-home} with the corresponding installation guide.

Two different prototype implementations are provided: \lpaas{} as a \emph{RESTful Web Service}, and \lpaas{} as an \emph{agent in an agent society} (MAS), both built on top of the \tuprolog{} system which provides the required interoperability and customisation.
The first aims to emphasise how \lpaas{} can effectively support REST, probably the most typical \iot{} paradigm, while the second means to highlight the \lpaas{} effectiveness in supporting and promoting distributed situated intelligence.

The concrete implementations are discussed in detail in \citeN{lpaas-icnsc2017}, \citeN{lpaas-ijgug2018}.
In \citeN{lpaas-icnsc2017}, a Smart Bathroom is supposed to monitor physiological functions to deduce symptoms and diseases, alerting the user via an Android app as appropriate: local sensors could perform situated reasoning, applying their local knowledge to aggregate the raw data and produce higher level synthesised information.
\citeN{lpaas-ijgug2018}, instead, discusses a Smart Kitchen where devices provide information about food supply and users' preferences, generating high-level knowledge used to coordinate and collaborate with other entities in the system.

\section{Case Study and Discussion}
\labelsec{discussion}

The following section is meant to illustrate the \lpaas{} approach and its benefits by means of a running example in the Smart House field: then, we compare the result with a more traditional LP approach.

The case study concerns the automatic assembly of home furniture by a domestic robot. The robot is in charge of the assembly operation, and the furniture pieces are supposed to be augmented with some form of computational capability---from simple RFID tags for being discoverable, up to embedded chips to store data and perform simple inference tasks.
The key aspect is that the installation instruction, the location, and the assembly constraints (such as avoid putting heavy things on fragile walls) are not known in advance by the robot itself, and have to be derived by suitably exploiting situated knowledge.

\begin{figure}
	\centering
  	\includegraphics[width=\linewidth]{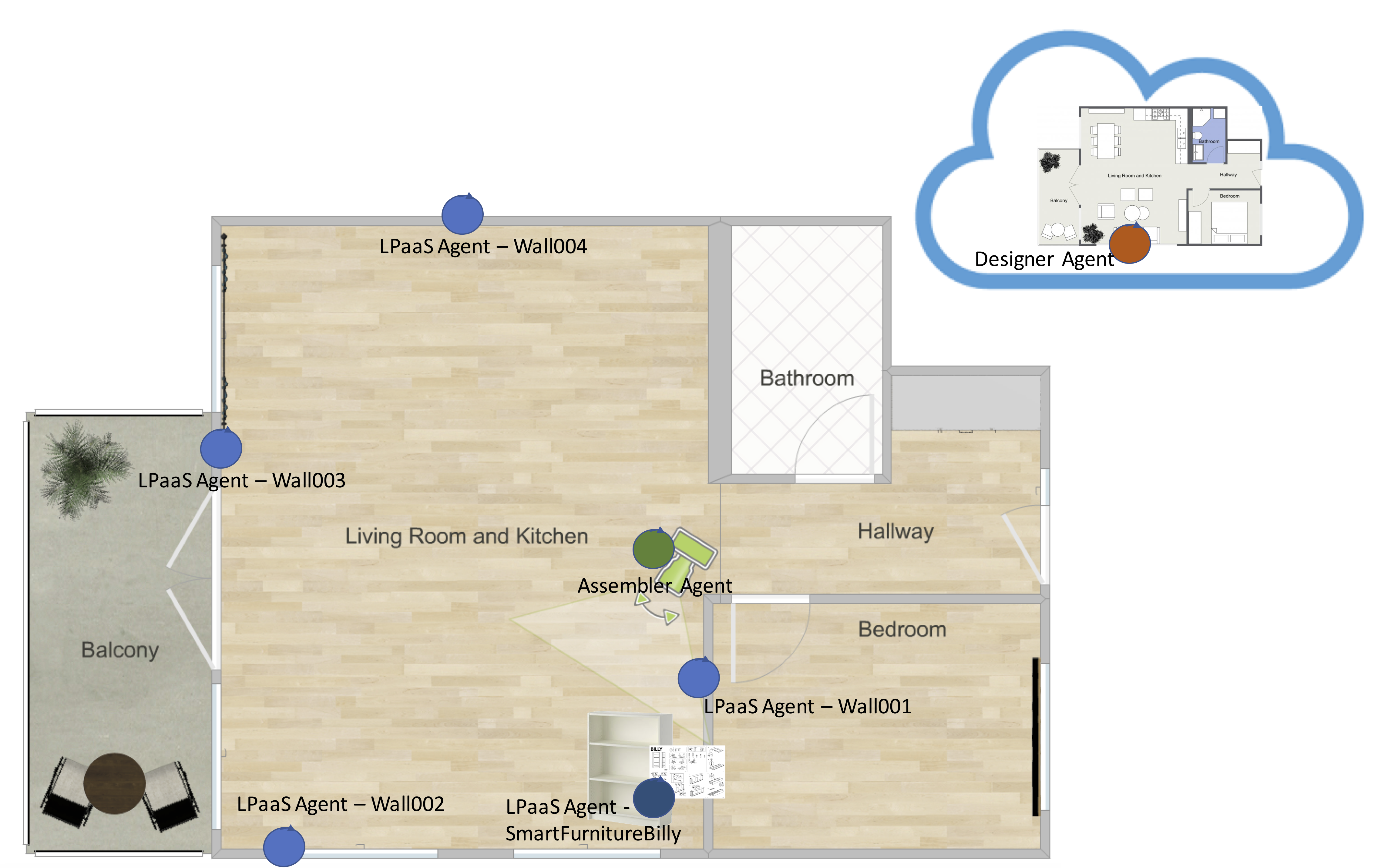}
  	\caption{\iot{} home: case study}\labelfig{plant}
\end{figure}

As shown in \xf{plant}, the envisioned system features the following actors:
\begin{itemize}
	\item the Designer agent, hosted in the Cloud, owner of and responsible for the house design project;
	\item the Assembler robotic agent, responsible for assembling the furniture;
	\item SmartFurniture \lpaas{} agents, owners of and responsible for storing and making available their own assembly instructions (i.e. the ``Billy'' bookshelf provides installation instructions for itself only);
	\item SmartWall \lpaas{} agents, each responsible for knowing the structural properties of a given wall (i.e.\ materials of constructions, maximum allowable weight, etc.).
\end{itemize}
The Assembler acts like a human with the goal of assembling all the furniture in the house according to the envisioned design, but conscious of the unexpected contingencies that may arise (fragile walls, wrong measures, etc.).
Moreover, exactly like a human, it does not know how to assemble a given piece of furniture in advance---it needs the installation instructions.
Accordingly, the Assembler first acquires the \emph{procedural knowledge} it needs -- essentially, the set of plans for assembling the furniture of interest -- by exploiting the intelligence embedded in the surrounding environmental structures (the walls, the ceiling, the floor) and furniture.
To this end, the Assembler needs not to be a full-fledged \lpaas{} agent, but can be a much simpler \lpaas{} client, with just the capability of requesting the \lpaas{} service.
Its (normal) workflow is thus as follows:
\begin{itemize}
	\item once in a room, it selects a wall and asks the Designer which pieces of furniture are to be positioned against that wall;
	\item then, it starts discovering the \lpaas{} agents representing such pieces, and asks them the pre-conditions they need for a successful assembly---i.e., about the structural properties the wall should have, or any other relevant property;
	\item then, the Assembler interacts with the targeted SmartWall \lpaas{} agent to check if such pre-conditions are satisfied---notice that this check is delegated to the wall itself, which is the only one bearing the situated knowledge needed to effectively evaluate the feasibility of the design project.
\end{itemize}
In case of unforeseen situations, the SmartWall agent proposes an alternative disposition of the furniture, that would be possibly implemented by the Assembler if the Planner agrees.
Yet again, the Assembler simply exploits the situated intelligence of the \lpaas{} services in its surroundings.
We would like to emphasise that this is the only way in which the same simple robot may be able to assemble (in principle) an unbounded number of heterogeneous pieces of furniture in all sorts of different walls, ceilings, floors, without hard-coding the instructions in its knowledge base, design update/patch mechanisms, or resort to code-on-demand features.

This is the kind of situated scenarios that \lpaas{} is most suited for, especially if compared with traditional LP approaches.
There, in fact, the robot would be in need of storing its own knowledge base -- the logic theory -- describing how to build the furniture, how to match the single pieces of furniture against walls, ceilings, etc., and overall the whole home project.
Besides leading to undesired centralisation and to a monolithic design, the most negative effect of such an approach is that the robot would be unable to work with new, unknown kinds of furniture---ultimately hindering flexibility, extensibility, and maintainability over time.

\begin{figure}
	\centering
	\includegraphics[width=0.85\linewidth]{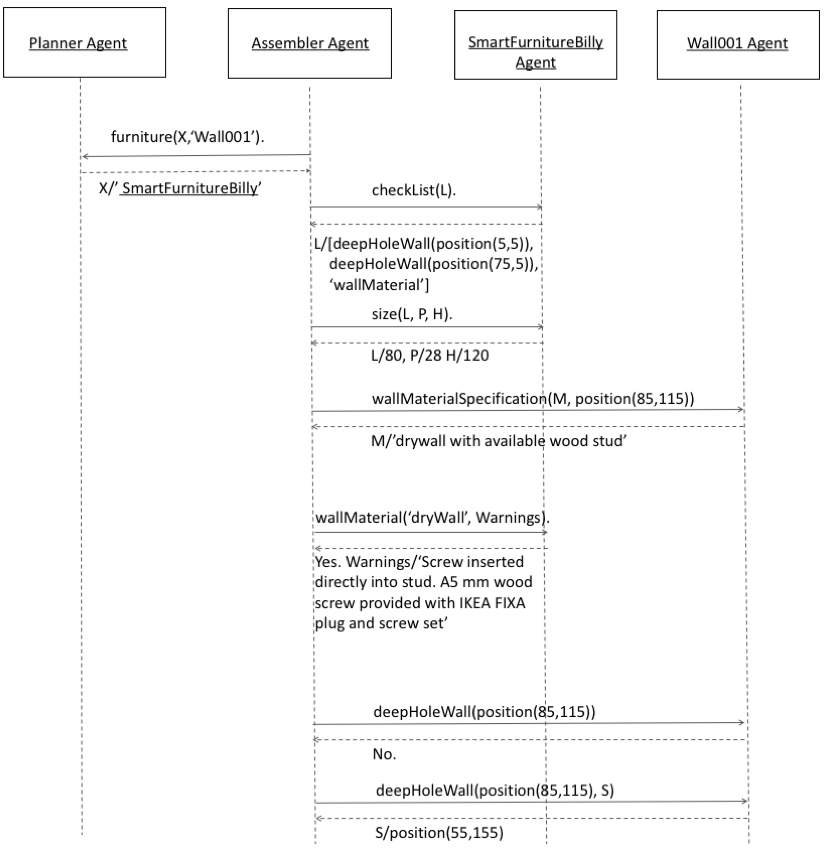}
  	\caption{Example of agents interaction}\labelfig{agentInteraction}
\end{figure}

An example run is shown in \xf{agentInteraction}:
\begin{itemize}
	\item fixing the library to the wall requires two deep holes, but one of them cannot be done due to a chimney (detected by the SmartWall agent)
	\item the SmartWall then, undertaking an inference step on its own knowledge base, proposes an alternative solution (i.e. to move the position of the hole of 30cm), which is then implemented by the Assembler after the Designer's approval.
\end{itemize}
It is worth noting that all the SmartFurniture and SmartWall agents implement, respectively, the same service -- that is, all SmartFurniture agents answer the same queries, and all SmartWall agents do the same for their own set of queries --; however, the answer, unlike traditional LP, could be different because of the surrounding situation.
For instance, given the query \texttt{wallMaterialSpecification(M, position(X,Y))}, agent \texttt{SmartWall001}, responsible for wall \#1, based on its wall material could reply \texttt{M/`drywall with available wood stud'}, while agent \texttt{SmartWall002}, responsible for a different wall with different characteristics, could reply \texttt{M/`masonry'}.

Despite its simplicity, the case study above highlights the effectiveness of the \lpaas{} approach in spreading intelligence in pervasive systems enabling ubiquitous intelligence. 
The approach turns out to be particularly interesting when dealing with different contexts, because taking into account local knowledge, situated in time and space, enables the system to take autonomous real time decisions based on the specific situation.
Moreover, relying mostly on locally available information reduces both the bandwidth consumption and the need for reliable communications between the distributed components, which are highly-desirable features in \iot{} scenarios.

Also, modularity and encapsulation of \lpaas{} improve \emph{scalability} of the \lpaas{} deployment: in traditional LP, there would be a single LP engine to scale up/down depending on the demand coming from clients, which translates to scaling a singleton monolithic entity as a whole, even if the actual need for scaling only concerns a portion of its functionality---i.e.\ only queries regarding a given portion of its knowledge base.
In \lpaas{} instead -- provided that the overall LP system functionality has been appropriately designed by distributing sub-functionalities to a set of distributed situated \lpaas{} services -- whenever any given portion of the LP inference service suffers from excessive demand, only that portion of the system needs to be scaled up---namely, only that \lpaas{} service instance.

Another benefit related to modularity and encapsulation is that \lpaas{} lends itself to application in real-time scenarios.
First of all, splitting out the LP service in multiple, smaller instances, responsible for a well-defined portion of the knowledge needed by the application at hand, helps achieving greater performance while doing inference, compared to traditional LP.
Second, time-awareness of \lpaas{} helps dealing with time-related aspects, such as discarding obsolete knowledge, ignore old requests, setting temporal bounds on the resolution process, etc.
Third, the \iot{}-oriented perspective according to which \lpaas{} clients do not need assert/retract mechanisms for manipulating the LP service knowledge base, because that functionality is envisioned for situated sensors and actuators directly interacting with the service through a dedicated API, helps ensuring that each LP service instance is always up-to-date with the state of the local world as per sensors' perceptions.

Nevertheless, we would like to point out that a real-time deployment of the currently available \lpaas{} prototypes -- the RESTful and MAS-oriented \tuprolog{} implementations -- has not yet been experimented, thus the above discussion is mostly based on speculation.

\section{Related Work}
\labelsec{related}

The SOA paradigm is widely used in \iot{} scenarios \cite{Messina2017,Karnouskos2012,Cannata2010,Guinard2010WoT,Guinard2010,pontelli2008}.
Moreover, communication via REST enables the direct integration of SOA-ready devices (i.e.\ devices hosting native web services).

MobIoT \cite{mobiot-pmc10} provides efficient service discovery, composition, and access in heterogeneous, dynamic, mobile \iot{} contexts, revisiting the standard SOA approach by providing probabilistic registration, look-up and thing-based composition based on comprehensive ontologies.
However, it does not support runtime interaction with users to let them specify their goals, and still needs proper validation from the scalability viewpoint when the number of registered services is very large.

\citeN{pontelli2008} present a comprehensive LP framework designed to support intelligent composition of Web services. 
The work proposes a theoretical framework for reasoning with heterogeneous knowledge bases, which can be combined with logic programming-based planners for Web service composition. 
The framework makes a step towards the interoperation between knowledge bases encoded using different rule markup languages and the integration of different components that reason about knowledge bases. Unlike our framework, the system is not focused on situated reasoning: rather, it is mainly concerned about dealing with heterogeneous Web services in the context of the WS composition problems.

A novel approach for engineering \iot{} systems is proposed by \citeN{Alkhabbas2017}, where a set of things with their functionalities and services is connected and led to cooperate temporarily so as to achieve a given goal.
Moreover, many research work deal with event-driven SOA (EDA-SOA) \cite{schulte2003event,michelson2006event} --- where communication between users, applications and services is carried out by events, rather than using remote procedure calls. 
In particular \citeN{Prado2017} propose an event-driven SOA which provides context awareness in the scope of \iot{}, whereby the generation of an event can trigger the concurrent execution of one or more services. 
When a given event occurs, different services can be triggered automatically, endowing the system with the capability of real-time sensing and rapid response to events in a loosely coupled, distributed computing environment. 
In general, pure SOA and EDA have their own limitations, but could complement each other; that is, some degree of service coordination can be achieved among mutually-independent services through the event mechanism. 
As mentioned by \citeN{Cheng2017}, such a complementarity suits well the features of \iot{}, requiring high autonomy inside a domain and efficient coordination across domains; furthermore, it both improves the real-time response to constantly changing business requirements and minimises the impact on the existing application system to allow a large-scale, distributed \iot{} service application to be easily developed and maintained.

Many aspects developed in the aforementioned works have worked as sources of inspiration for \lpaas{}, in particular in how the model and architecture are conceived and designed.
Following the SOA principles, \lpaas{} aims at modelling ubiquitous intelligence in a dynamic context by promoting portability and interoperable interaction over a network (via proper standards) and emphasising the separation of the service interface from its implementation.
While following the EDA principles, \lpaas{} goes beyond the state-of-the-art mainly as far as context-awareness is concerned, in particular by supporting the injection of intelligence within existing services/agents via the awareness of the context, thus promoting their adaptivity.

\section{Conclusion \& Future Work}
\labelsec{conclusion}

In this paper we propose the \lpaas{} approach for \emph{distributed situated intelligence} as the natural evolution of LP in the context of nowadays pervasive computing systems.
We discuss its properties and its computational and architectural models by relating and comparing them to the notions and development of LP over the years, tracked in \xs{related}.

The main advantages of exploiting an LP-based approach in pervasive systems amount at
\textit{(i)} writing declaratively complex rules involving the context, 
\textit{(ii)} assessing provable statements about the expressive power and decidability of the context model, and 
\textit{(iii)} actually supporting light-weight reasoning and cooperation among distributed components.
We also present a first prototype implementation built on top of the \tuprolog{} system, to demonstrate and test the effectiveness of the \lpaas{} approach.

Our service-based approach, in particular, 
\emph{(i)} encourages representing and reasoning with situations using a declarative language, providing a high level of abstraction; 
\emph{(ii)} supports the incremental construction of context-aware systems by providing modularity and separation of concerns; 
\emph{(iii)} promotes the cooperation and interoperation among the different entities of a pervasive system; and 
\emph{(iv)} enables reasoning over data streams, like those collected by sensors.

Of course, a number of enhancements are still possible, both to the model and to the infrastructure.
From the model viewpoint, specific \emph{space-awareness} methods could be defined and added: for instance, a \texttt{solveNeighbours} primitive to deal with the space around either the client or the server, making it possible to opportunistically federate LP engines upon need as a form of dynamic service composition.
From the infrastructure viewpoint, we plan to focus on the design and implementation of a specialised LP-oriented middleware, dealing with heterogeneity of platforms as well as with distribution, life-cycle, interoperability, and coordination of multiple situated \prolog{} engines -- possibly based on the existing \tuprolog{} technology  and \tucson{} middleware \cite{tucson-jaamas2} -- so as to explore the full potential of logic-based technologies in \iot{} scenarios and applications.
Also, providing some sort of distributed service directories enabling dynamic discovery of \lpaas{} services -- in turn promoting opportunistic interactions with clients or other services (for service composition) -- is surely a promising path to follow to further widen the applicability of the \lpaas{} approach in pervasive scenarios.

\bibliographystyle{acmtrans}
\bibliography{CDMO-DLP-TPLP2017}
\normalsize
\end{document}